\documentclass[]{article}

\usepackage{authblk}
\usepackage[T1]{fontenc}
\usepackage[utf8]{inputenc}
\usepackage{amsmath,amssymb,amsfonts}
\usepackage{algorithm,algpseudocode}
\usepackage{url}
\usepackage{multirow}
\usepackage{tikz}

\DeclareMathOperator*{\argmax}{arg\,max}
\DeclareMathOperator*{\argmin}{arg\,min}
\newcommand{\norm}[1]{\left\lVert#1\right\rVert}

\newcommand{\mbs}[1]{\boldsymbol{#1}}

\title{Tuning metaheuristics by sequential optimization of regression models.\footnote{Submitted for publication in \textit{Swarm Intelligence} on September 10 2018. Returned by the editor-in-chief for journal scope reasons on September 14 2018. Submitted to \textit{Information Sciences} on September 21, 2018. This work has been partially funded by Brazilian research agencies CNPq (grant: 404988/2016-4), CAPES, and Fapemig (grant: APQ-01099-16).}}
\author[1,2]{\'Athila R. Trindade}
\author[1,3]{Felipe Campelo}
\affil[1]{Operations Research and Complex Systems Laboratory\\
  Universidade Federal de Minas Gerais\\
  Belo Horizonte 31270-010, Brazil.}
\affil[2]{Graduate Program in Electrical Engineering\\
  Universidade Federal de Minas Gerais, Brazil.
  E-mail: athilart@ufmg.br}
\affil[3]{Department of Electrical Engineering\\
  Universidade Federal de Minas Gerais, Brazil.
  E-mail: fcampelo@ufmg.br}
  
\begin{document}
\date{September 21, 2018}
\maketitle

\begin{abstract}
Tuning parameters is an important step for the application of metaheuristics to problem classes of interest. In this work we present a tuning framework based on the sequential optimization of perturbed regression models. Besides providing algorithm configurations with good expected performance, the proposed methodology can also provide insights on the relevance of each parameter and their interactions, as well as models of expected algorithm performance for a given problem class, conditional on the parameter values. A test case is presented for the tuning of six parameters of a decomposition-based multiobjective optimization algorithm, in which an instantiation of the proposed framework is compared against the results obtained by the most recent version the Iterated Racing (Irace) procedure. The results suggest that the proposed approach returns solutions that are as good as those of Irace in terms of mean performance, with the advantage of providing more information on the relevance and effect of each parameter on the expected performance of the algorithm.
\end{abstract}

\section{Introduction}
\label{introd}
Metaheuristics such as evolutionary algorithms \cite{Gendreau2010,Eiben2003} represent a class of computational problem solvers subject to stochastic behavior, determined in part by the values of user-defined parameters. These parameters are responsible for determining the global-local exploration profile, solution quality, and efficiency of the algorithm when searching for solutions in the objective space. Poor choices of parameter values can result in low performance of the method, even if the implementation is done properly, while well-chosen values can lead the algorithm to consistently return high-quality solutions. Moreover, good parameter configurations are often problem-dependent \cite{Birattari2005}, which limits the utility of looking for one-size-fits-all configurations and requires the development of efficient strategies for tuning parameters based on a limited sample of representative instances of the problem class of interest.

Assuming that parameters can assume several (sometimes infinitely many) values, a possibly very large number of combinations of parameter values -- called here \textit{candidate configurations}, or simply \textit{configurations} -- can be considered for an algorithm when solving a given problem. There are two sources of random variation in the expected performance of an algorithm (equipped with a given configuration) when solving instances of a given problem class: the uncertainty due to the instance being solved, which gives rise to an \textit{across-instances variance}; and the uncertainty due to the stochastic behavior of the metaheuristic itself, which results in a \textit{within-instance variance} \cite{Birattari2005}. Due to these random influences on the observed performance of a given algorithm configuration, several researchers have proposed strategies for recommending candidate configurations based on statistical concepts, in a process commonly referred to as \textit{parameter tuning} \cite{Eiben2011,Hoos2012}.

This work is focused on the application of statistical modeling to the development of tuning approaches. More specifically, we present a modular framework for implementing parameter tuning methods, which is based on concepts drawn from Sequential Model Based Optimization (SMBO) strategies \cite{Bartz-Beielstein2009,Jones1998}. The proposed framework is aimed not only at returning algorithm configurations that are well-adjusted for particular problem classes, but also to provide statistical models capable of supporting further investigations on the relative relevance of algorithm parameters and interaction effects, as well as estimations of expected algorithm performance given new sets of parameter values.

The remainder of this manuscript is organized as follows. We start by formally stating the \textit{Algorithm Tuning Problem} that we are attempting to solve (Section \ref{algconfprob}), and briefly reviewing the most widely used parameter tuning methods (Section \ref{paramtumeth}). The proposed tuning framework is introduced in Section \ref{proptunfram}. To illustrate its use, we consider the problem of tuning six parameters from an evolutionary multiobjective optimization algorithm, and contrast the results obtained by the proposed method against those returned by Iterated Racing \cite{Lopez-Ibanez2016}. Finally, some conclusions and possibilities of continuity are explored in Section \ref{sec:conclusions}.

\section{The Algorithm Tuning Problem}
\label{algconfprob}

In this work we are interested in tuning algorithm parameters for a given problem class of interest, i.e., finding the combination of parameter values that results in the best expected performance of a given algorithm on instances belonging to a given family of problems. Here we present a formalization of this problem, based on a description originally presented by Birattari \cite{Birattari2004,Birattari2005}.

Assume that we have an algorithm containing $k$ free parameters to be set by the user, and let $\theta_i$ denote a list of length $k$ containing specific parameter values for that algorithm. We refer to $\theta_i$ as a candidate configuration for the algorithm under study, with $\Theta = \left\{\theta_1,\theta_2,\dotsc \right\}$ representing the set of all possible parameter configurations for that algorithm.\footnote{For the sake of simplicity, in the remainder of this work we refer to \textit{the algorithm equipped with a given set of parameter values} as simply \textit{a configuration}.} Similarly, let $\gamma_j$ denote a given problem instance belonging to a problem class of interest, denoted by $\Gamma = \left\{\gamma_1,\gamma_2,\dotsc\right\}$. Also, let $X_{ij}$ be a random variable representing the performance\footnote{Measured according to a given indicator of choice. In the remainder of this work, we assume the use of indicators for which larger values represent better performance.} of a candidate configuration $\theta_i\in\Theta$ on a given instance $\gamma_j\in\Gamma$, with $\varphi_{ij}$ denoting a statistical parameter of $X_{ij}$ that can be used to quantify the general quality of configuration $\theta_i$ as a solver of instance $\gamma_j$, e.g., the mean or median of $X_{ij}$.

Let $\Phi_{i:\Gamma} = \left\{\varphi_{ij}: \gamma_j\in\Gamma\right\}$ denote the set of quality values of a candidate configuration $\theta_i$ for all instances belonging to problem class $\Gamma$; and $\mu_{i:\Gamma}$ denote a statistical parameter of $\Phi_{i:\Gamma}$ which is of interest when comparing different configurations, e.g., the mean or the median performance across all instances belonging to $\Gamma$. Under these definitions, the algorithm tuning problem tackled in this work can be defined as:%
\begin{equation}\label{eq:probdef}
	\text{Find } \theta^* = \argmax_{\theta\in\Theta}~~\mu_{i:\Gamma},
\end{equation}

\noindent that is, the problem of finding the configuration that maximizes the performance of a given algorithm for a given class of instances. Automated approaches for addressing this problem generally try to obtain $\theta^*$ using information from a finite subset of problem instances. 

An important point to be aware of is that the instances used for tuning are usually not the ones that are relevant in practice: an underlying assumption of methods that attempt to solve the problem defined above is that the instances used for tuning can be regarded as a representative sample of the problem class of interest, and can therefore be used for modeling and inference of the expected behavior of the algorithm for that problem class. 

It must also be highlighted that, under this definition of the algorithm tuning problem, the \textit{independent observations} to be used in any statistical modeling or inferential procedure refer to individual estimates of performance of a given configuration on a given instance, i.e., to individual values of $\varphi_{ij}$. Repeated runs of a given configuration on the same instance are useful for improving the accuracy of estimates of these performance values, but cannot count as independent degrees-of-freedom for the statistical procedures. Failure to account for this particular fact would result in pseudoreplication \cite{Hurlbert1984,Lazic2010}, a violation of the assumption of independence underlying the statistical approaches used in most tuning procedures that leads to inflated type-I errors in inferential tests, and to artificially reduced standard errors in descriptive models.

In the next section we review some of the most common approaches used to tackle the algorithm configuration problem. While in most cases the problem is not explicitly stated as above, the workings of these methods indicate that in most cases this is the problem (or at least one of the problems) they attempt to solve. After briefly discussing the existing approaches, we will present our proposed tuning framework in Section \ref{proptunfram}.

\section{Overview of Parameter Tuning Methods}
\label{paramtumeth}

A variety of different tuning methods have been proposed over the years to determine the best configurations of algorithms when solving a given problem. Based on their working mechanisms and design principles, it is possible to group these methods in three major categories: racing methods, SMBO methods, and hyper-heuristics. In this section we review the most widely used methods from each category.

\subsection{Racing Methods}
\label{racing}

The basic concepts of racing methods were initially proposed in the machine learning literature for solving the \textit{model selection problem} \cite{Birattari2005}. The basic idea of these methods \cite{Maron1994,Moore1994} is that the search for the best model structure can be sped up by discarding inferior candidate models as soon as sufficient statistical evidence is gathered against them. A similar concept is used by racing methods for parameter tuning: discard candidate configurations as soon as they are detected as inferior according to some statistical criteria. 

The most relevant methods in this class are all based on concepts originally introduced in the form of the F-Race \cite{Birattari2002}. The main concept behind F-Race is to iteratively evaluate a given set of candidate configurations on a finite number of instances, gradually building statistical evidence until it is possible to conclude, at a predefined level of confidence, that one or more candidate configurations are significantly worse than the others. Once this is determined, those inferior configurations are eliminated and the process continues with the remaining ones. F-Race stops when a given termination criterion is observed, e.g., the maximum computational budget is used or the number of remaining configurations falls below a given threshold. At each iteration this method employs Friedman tests \cite{Sheskin2011} as their main inferential procedure, followed (if statistically significant differences are detected) by post-hoc nonparametric pairwise comparisons between the estimated best configuration and all others. Configurations whose median performance is detected as significantly worse than that of the best one are discarded from the race. The F-Race method then proceeds by evaluating the remaining configurations on more instances, iteratively increasing the statistical power of the tests and enabling the detection of smaller differences in median performance. The method stops when only a single configuration remains, a given number of instances have been sampled, or a predefined computational budget has been exhausted. 

Improvements to F-Race were proposed in the form of the Iterated F-Race (I/F-Race) method \cite{Balaprakash2007}, later generalized as Iterated Racing \cite{Lopez-Ibanez2016}. I/F-Race works by iteratively applying F-Race, generating new candidate configurations at each iteration by sampling from a multivariate random distribution of parameter values that is biased by the best configurations returned in the previous iterations \cite{Birattari2010}. This biased sampling drives the search process towards obtaining candidate configurations that are similar to the best ones observed up to a given iteration. \textit{Iterated Racing} allows the use of different statistical tests in place of the Friedman test, prevents premature convergence of the tuning method by means of soft restart rules, and include elitist options to force the preservation of high-quality candidate configurations.

\subsection{SMBO Methods}
\label{smbo}

Tuning methods based on the sequential model-based optimization (SMBO) approach are motivated by results from the literature on statistical modeling and black-box optimization methods. From an initial set of observations of performance over the space of configurations, SMBO methods fit one or more response surfaces, which are then used to determine which new configurations should be sampled. These new results are then added to the existing sample, and used to update the response surfaces. As iterations progress, SMBO methods tend to generate models that are increasingly biased towards those regions of the parameter space which contain configurations with good performance. The three most widely known tuning methods based on SMBO are Sequential Parameter Optimization (SPO) \cite{Bartz-Beielstein2005}, BONESA \cite{Eiben2011}, and Sequential Model-Based Algorithm Configuration (SMAC) \cite{Hutter2011}, which are briefly discussed below.

Sequential Parameter Optimization was proposed in 2005 \cite{Bartz-Beielstein2005,package-spot}, and is based on a strategy of iteratively improving a prediction model to reveal the relationship between parameter values and algorithm performance. This model is then used to select the most promising values for the parameters. In the first iteration of SPOT a few candidate configurations are generated using Latin Hypercube Sampling (LHS) \cite{McKay1979,Wyss1998} over the space of algorithm parameters. These candidate configurations are evaluated on a problem instance, and this information is used to fit a statistical prediction model. The standard initial model used by SPOT is a second-order linear regression model, but regression trees and Kriging have also been employed \cite{Bartz-Beielstein2009}. Based on the candidate configuration with the best observed performance and on the response surface, new candidate configurations are generated so as to maximize the probability, conditional on the available information, that they will present good performance values. These new points are evaluated and an updated model is fit, in a process that iterates until a predefined termination criterion is reached. At each cycle, the number of evaluations of each candidate configuration on the problem instance is increased, obtaining more accurate estimations of average performance. Besides searching for the best configuration, SPOT also allows the user to analyze the variation of algorithm behavior with its parameter values using the statistical models generated, thereby enabling deeper experimental investigations and experiment-driven algorithm development. 

BONESA \cite{Eiben2011} is a tuning method based on \textit{learning} and \textit{searching} loops. These two modules continuously exchange information as iterations progress: the learning loop uses a prediction model to compare candidate configurations, while the searching loop is responsible for sampling new candidate configurations based on the results of the learning module. The distinguishing feature of this method is its multi-objective approach: to select the best parameter values for a given problem class, BONESA uses a Pareto strength approach \cite{Eiben2011} and attempts to simultaneously maximize the performance of the algorithm for all problem instances used in the tuning effort.

In the first iteration, BONESA randomly samples a number of candidate configurations, evaluating them once for each available tuning instance. The learning loop uses this information to predict the utility values for new candidate configurations, using an approach based on the weighted average of the utilities of the nearest neighbors of the proposed configurations. These predicted utilities are then used for comparing the candidate configurations using a criterion based on Pareto dominance and an adaptation of Welch's t test \cite{Montgomery2012}. The results of the tests are then aggregated and used to calculate the Pareto strength of each candidate configuration \cite{Eiben2011} and to generate new configurations (based on the best ones), for which the Pareto strength is also calculated. Then, those with the highest Pareto strength values are selected to compose the new set of configurations to be evaluated on the tuning instances. The method iterates until a given stop criterion is reached.

Finally, the Sequential Model-Based Algorithm Configuration (SMAC) method \cite{Hutter2011,Hutter2013} was, similarly to the SPO, initially designed for tuning algorithm parameters on a single problem instance.\footnote{Both methods can, however, be adapted for tuning algorithms for problem classes.} The method generates an initial set of candidate configurations and evaluates their performance on the instance. Based on this information, it fits a predictive model of performance over the space of parameter values, and then performs a multi-start search for finding the candidate configuration that maximizes an expected positive improvement function. This new candidate configuration is then evaluated and added to the pool of candidate configurations, and the process is repeated. SMAC has been used with different types of prediction model, including Gaussian Processes and Random Forests; and different search strategies, including DIRECT and CMA-ES. 

%Computational experiments presented by Hutter \textit{et al.} \cite{Hutter2011,Hutter2013} suggest that SMAC is a robust tuning method when compared against SPO and ParamILS (see Section \ref{paramils}) for the tuning of three solvers, namely SAPS (a dynamic local search algorithm for SAT problems), SPEAR (another solver for SAT problems) and CPLEX 10.1.1 (a commercial solver for mixed integer programming problems) for single and multiple instances.

\subsection{Hyperheuristics}
\label{hiperheu}

The term \textit{hyperheuristics} is used here to classify those tuning methods which consist in the application of metaheuristics for obtaining the best parameter values of algorithms, trying to solve the algorithm configuration problem by directly tackling its optimization formulation, discussed in Section \ref{algconfprob}. While in principle any optimization approach could be used to solve the algorithm tuning problem, knowledge about the characteristics of this problem have motivated the development of specific strategies. Two of the most common ones are REVAC \cite{Nannen2006,Nannen2007} and ParamILS \cite{Hutter2009}, as presented below.

Nannen and Eiben proposed a parameter tuning method for Evolutionary Algorithms called \textit{Relevance Estimation and Value Calibration} (REVAC) \cite{Nannen2006,Nannen2007}, which aims to answer questions related to two aspects of algorithm design and configuration: (i) which of the free parameters of a given method are in fact relevant, i.e., effectively influence the performance of the algorithm; (ii) for those parameters that are in fact relevant, which values lead to the best performance of the algorithm. 

REVAC is itself configured as an evolutionary strategy. The method begins with a population of randomly generated candidate configurations, which are evaluated according to a performance function, and new candidate configurations are obtained using usual recombination and mutation operators \cite{Nannen2007}. At each iteration the marginal probability density functions for each parameter of the algorithm are estimated from the population of candidate configurations. The Shannon entropy of these distributions is used to estimate the relevance of each parameter. Parameters for which entropy decreases quickly as iterations progress need little information to be tuned, and are therefore considered more relevant to the performance of the EA. Conversely, those for which entropy does not decrease are considered less relevant, and may be discarded or receive arbitrary values. The method iterates until predefined stop criteria are reached.

ParamILS \cite{Hutter2009} is a framework of tuning methods, which is based on Iterated Local Search (ILS). Starting from a given initial candidate configuration, at each iteration the incumbent configuration is perturbed and undergoes a first improvement local search, to generate a new candidate configuration that replaces the incumbent one if it presents better performance. The neighborhood of a given configuration is the set of all configurations that differ from it in a single parameter, and the determination of whether a candidate configuration is better than the incumbent one is performed using statistical tests, with problem instances as a blocking factor \cite{Hutter2009,Montgomery2012}. Variants of this basic algorithm include \cite{Hutter2009} \textit{FocusedILS}, which adaptively selects the number of training instances; and \textit{Adaptive Capping of Algorithm Runs}, which controls the cutoff time for each run of the candidate configurations. %These approaches, together with the basic one, were tested for tuning the same 3 solvers used for testing SMAC, namely SAPS, SPEAR and CPLEX, for three distinct sets of test problems. In all cases the results were superior to those obtained using default configurations.

%\subsection{Performance of parameter tuning methods}
%\label{comptunmeth}
%
%A comparison of 5 tuning methods, namely REVAC, SPO, BONESA, SMAC and Iterated F-Race, was presented in 2012 by Selmar Smit \cite{Smit2012}. The target algorithm chosen to be tuned was the $(1,\lambda^s_m)$-CMA-ES \cite{Auger2010}, for which six numeric (integer and real) parameters were selected for tuning. Five tuning instances were used in the experiment, and the performances of the best parameter values found by the tuning methods were compared with those obtained by the standard parameter values used in the BBOB'10 competition \cite{Hansen2010}. Performance was measured as the number of function evaluations needed to reach a certain quality threshold, with a maximum allowed number of evaluations of 20000.
%
%Ten runs of each tuning method on each test problem were performed, and the resulting configurations were ran 100 times on the test problems, with the average performance recorded. Considering each problem, BONESA and SMAC obtained the best results both on performance and stability. Another interesting result was that for two problems none of the tuners managed to find a parameter vector that performed significantly better than the recommended parameter values \cite{Auger2010} used for the BBOB'10 competition. The only approach tested for the case of a multi-instance tuning, BONESA, was able to identify a candidate configuration which was better than the recommended parameter values in a half of the runs on all problem instances.
%

\section{Proposed Tuning Framework}
\label{proptunfram}

In this section we propose a modular structure for tackling the algorithm configuration problem presented in Section \ref{algconfprob}. The proposed framework, which we will refer to as \textit{MetaTuner}, can be used to instantiate distinct tuning approaches through the adoption of specific methods for each of its components, depending on the nature of the tuning process at hand. This modular approach results not only in a greater flexibility for the framework, but is also useful for faster development and testing of proposed improvements.

The proposed approach is based on a common assumption in the design of computer experiments \cite{Sacks1989}, that if the number of instances and of candidate configurations is sufficient, enough information will be gathered so that the resulting response surfaces are somehow representative of the expected performance landscape of the algorithm for the problem class of interest. Under this assumption, optimizing these surfaces will tend to drive the method towards regions of the parameter space containing good candidate configurations, allowing the method to iteratively concentrate its efforts on those regions of the parameter space with the highest average performance. 

\begin{algorithm}
	\caption{Proposed tuning framework}
	\label{algo:metatun}
	\begin{algorithmic}[1]
		\Require Search space for parameters ($\Omega$); Set of tuning instances ($\Gamma_S$); number of initial configurations ($m_0$); number of new configurations per iteration ($m_\star$); number of initial instances ($N_0$); number of additional instances per iteration ($N_\star$); size of elite archive ($n_\mathcal{E}$); initial sampler parameters; stop criteria parameters; regression model parameters; optimizer parameters.
		\State $t\leftarrow 0$
		\State $\mathcal{A}^{(t)} \leftarrow$ GenerateInitialSample $\left(\Omega,m_0\right)$ \Comment{Sample initial configurations}\label{alg:initcand}
		\State $\Gamma_{\mathcal{A}}^{(t)} \leftarrow$ SampleWithoutReplacement $\left(\Gamma_S,N_{0}-N_{\star}\right)$ \Comment{Sample initial instances}
		\State $\mathcal{P}_\mathcal{A}^{(t)} \leftarrow$ EvaluateConfigurations $\left(\mathcal{A}^{(t)},\Gamma_{A}^{(t)}\right)$ \Comment{Evaluate configurations on instances} \label{alg:evalcand1}
		\State $\mathcal{E}^{(t)} \leftarrow \mathcal{A}^{(t)}$ \Comment{Initialize elite archive}
		\While{Stop criteria not met}
		\State $t\leftarrow t+1$
		\State $\Gamma^{\prime} \leftarrow$ SampleWithoutReplacement $\left(\Gamma_S\backslash\Gamma_{\mathcal{A}}^{(t-1)},N_{\star}\right)$ \Comment{Sample $N_{\star}$ new instances}
		\State $\mathcal{P}^\prime \leftarrow$ EvaluateConfigurations $\left(\mathcal{E}^{(t)},\Gamma^\prime\right)$\Comment{Evaluate elite configs. on new instances}\label{alg:evalcand2}
		\State $\Gamma_{\mathcal{A}}^{(t)}\leftarrow\Gamma_{\mathcal{A}}^{(t-1)}\bigcup\Gamma^\prime$\Comment{Update archive of instances visited}
		\State $\mathcal{P}_{\mathcal{A}}^{(t)}\leftarrow$ Update $\left(\mathcal{P}_{\mathcal{A}}^{(t-1)},\mathcal{P}^\prime\right)$\Comment{Update archive of config. performances}
		\State $\mathcal{S}_1^{(t)}\leftarrow$ FitModel$\left(\mathcal{A}^{(t-1)},\mathcal{P}_{\mathcal{A}}^{(t)}\right)$\Comment{Fit regression model}
		\State $\theta_1^{\prime{(t)}}\leftarrow$ Optimize $\left(S_1^{(t)}\right)$\Comment{Find configuration that optimizes $\mathcal{S}_1^{(t)}$}\label{alg:optim1}
		\For{$j\in\left\{2,\dotsc,m_\star\right\}$}
		\State $\mathcal{S}_j^{(t)}\leftarrow$ PerturbModel $\left(S_1^{(t)}\right)$\Comment{Generate perturbed model}\label{alg:perturbmodels}
		\State $\theta_j^{\prime{(t)}}\leftarrow$ Optimize $\left(S_j^{(t)}\right)$\Comment{Find configuration that optimizes $\mathcal{S}_j^{(t)}$}
		\EndFor \label{alg:optim2}
		\State $\mathcal{C}^{(t)}\leftarrow\left\{\theta_j^{\prime{(t)}}:j=1,\dotsc,m_\star\right\}$\label{alg:optim3}
		\State $\mathcal{P}_\mathcal{C}^{(t)} \leftarrow$ EvaluateConfigurations $\left(\mathcal{C}^{(t)},\Gamma_\mathcal{A}^{(t)}\right)$\Comment{Evaluate candidate configurations}\label{alg:evalcand3}
		\State $\mathcal{A}^{(t)}\leftarrow\mathcal{A}^{(t-1)}\bigcup\mathcal{C}^{(t)}$\Comment{Add candidate configs. to archive}\label{alg:arch1}
		\State $\mathcal{P}_{\mathcal{A}}^{(t)}\leftarrow$ Update $\left(\mathcal{P}_{\mathcal{A}}^{(t)},\mathcal{P}_{\mathcal{C}}^{(t)}\right)$\Comment{Update archive of config. performances}\label{alg:arch2}
		\State $\mathcal{E}^{(t)}\leftarrow$ SelectKBest $\left(\mathcal{A}^{(t)},\mathcal{P}_\mathcal{A}^{(t)},K = n_{\mathcal{E}}\right)$\Comment{Update elite archive}
		\EndWhile
		\State $\mathcal{P}_{\mathcal{E}}^{(t)}\leftarrow$ RetrieveElitePerformances$\left(\mathcal{E}^{(t)}, P_\mathcal{A}^{(t)}\right)$
		\vspace{.10cm}
		\State\Return Archive of elite configurations ($\mathcal{E}^{(t)}$), estimated performance of elite configurations ($\mathcal{P}_\mathcal{E}^{(t)}$); other statistics of interest.
	\end{algorithmic}
\end{algorithm}

The general aspects of the proposed framework can be easily explained from the structure presented in Algorithm \ref{algo:metatun}.\footnote{An open-source implementation is available in the form of \texttt{R} package \texttt{MetaTuner}:\\ \url{https://github.com/fcampelo/MetaTuner}} The method starts by sampling a few configurations, which are evaluated on a randomly sampled initial set of tuning instances. The performance results obtained are then used for fitting a regression model of the expected performance of configurations on the problem class of interest. The regression model is then subject to perturbations (e.g., by perturbing the fitted parameters), resulting in a number of additional response surfaces. For each surface (including the unperturbed one) an optimization process is executed, returning a new candidate configuration which maximizes the value of the estimated average performance value for that model. These new candidate configurations are then evaluated on all instances sampled so far, and added to an archive. Finally, the archive is truncated to a given size, maintaining only the candidate solutions with the best expected performance value for the problem class. The whole process then iterates by sampling a few more instances (if available), and proceeds until a predefined stopping condition is reached.

%Following the taxonomy presented in the previous section, the proposed approach can be seen as an SMBO method. The method iteratively samples and evaluates new configurations, builds predictive models of the average performance of algorithm configurations on the problem class of interest, and optimizes surface responses obtained by perturbing the predictive models, using the uncertainty associated with the estimation of their coefficients. The method draws some inspiration from racing procedures, particularly Iterated Racing, in that it sequentially discards configurations perceived as leading to worse expected performance and focuses on evaluating those that yield better results. The information gathered on those ``inferior'' configurations is not discarded, however, as it can provide important information for the model-building effort. 

In the remainder of this section we detail the implementation of an initial instantiation of the proposed framework, aimed at tuning continuous parameters.\footnote{The tuning of categorical or hierarchical parameters is not considered in the present work.}

\subsection{Generation of Initial Candidate Configurations}
\label{initcand}
Considering the importance of gathering enough information for generating a reasonable first set of regression models, a point of particular importance is to ensure that the sampling of initial candidates (line \ref{alg:initcand} of Algorithm \ref{algo:metatun}) be well-distributed in the parameter space, so that the method will have the chance to investigate different regions of the space of parameters. 

There are a few strategies that guarantee a well-spread initial sampling in continuous spaces. Some of the most widely known include Latin Hypercube Sampling (LHS) \cite{McKay1979,Ye1998}, low-discrepancy sequences of points (LDSP) \cite{Kuipers2005}, and uniform designs (UD) \cite{Ning2011}. Since LHS is possibly the one most widely used in computational experiments \cite{Santner2003}, the version of MetaTuner described here uses this particular sampling scheme for generating its initial set of candidate configurations.

Before proceeding, it is important to understand that performance degradation can occur if the parameters being tuned can assume values on possibly very different scales -- e.g., in the case of polynomial mutation \cite{Deb1999}, the rate parameter exists in the $\left[0,1\right]$ interval, while $\eta$ can in principle assume any non-negative value. This is a well-known issue in the regression and machine learning literature \cite{Witten2016}, which can be avoided when tuning numerical parameters by simply rescaling all parameters to a common scale, e.g., $[0,1]$:
\begin{equation}
\theta_{i(l)}^\prime = \frac{\theta_{i(l)}-\theta_{(l,min)}}{\theta_{(l,max)} - \theta_{(l,min)}},~~~i=1,\dotsc,m;
\label{eq:normalize2}
\end{equation}

\noindent where $\theta_{i(l)}$ is the value of the $l$-th component of candidate configuration $\theta_i$, and $\theta_{(l,min)}~\theta_{(l,max)}$ denote the lower and upper allowed values for the $l$-th parameter being tuned. Notice that this require all parameters to have upper and lower limits, which is generally not a problem -- even for parameters that are theoretically unbounded, it is generally possible to define reasonable bounds based on theory or previous experience.

\subsection{Evaluation of Candidate Configurations and Estimation of Quality Value}
\label{evalcand}
Given the possibly heterogeneous nature of the tuning instances and the expected variations of performance of different configurations, it is possible that the distributions of $X_{ij}$, i.e., of the performance of candidate configurations on the instances, exist on very different scales. While some regression models, particularly quantile regression \cite{Koenker2001}, can deal with these differences of scale relatively well, most have their performance heavily degraded in the presence of such large scale differences and heterogeneity of variances. To alleviate these particular problems, the performance of candidate configurations on the tuning instances (lines \ref{alg:evalcand1}, \ref{alg:evalcand2} and \ref{alg:evalcand3} of the algorithm) is calculated by running the configurations on the test instances and transforming the output (i.e., the value of the quality indicator used) to a common scale.

Let $x_{ij}\sim X_{ij}$ denote a single observation of the performance of configuration $\theta_i$ on instance $\gamma_j$. The observed performance in this case is calculated by linearly scaling $x_{ij}$ to the interval $\left[0,1\right]$:
\begin{equation}
x_{ij}^\prime = \frac{x_{ij} - x_{min,j}}{x_{max,j} - x_{min,j}},
\label{eq:normalize1}
\end{equation}

\noindent where $x_{min,j},~x_{max,j}$ denote the smallest and largest values observed so far for instance $j$, across all configurations already evaluated. Once these values are calculated for all instances visited by a given configuration $\theta_i$, the summary performance estimator $p_{\theta_i}$ is calculated as the sample average of the $x_{ij}^\prime$ values associated with that configuration. Notice that this average can be the simple mean, trimmed mean, median, or any other indicator of location. In the version used here, the median is employed due to its robustness to outliers and distributional asymmetries, an important characteristic when dealing with possibly heterogeneous tuning instances. 

Notice from Algorithm \ref{algo:metatun} that, at each iteration, configurations in the elite archive $\mathcal{E}^{(t)}$ are evaluated on the $N_{\star}$ new instances, while configurations that were not selected are not, even though all are used for modeling the average behavior. This is done to increase the accuracy of estimation of the average performance on the most promising configurations, and can be used, for instance, to attribute weights to each observation in the regression modeling. At each iteration, the new configurations generated by optimizing the estimated response surfaces, are also evaluated on all instances visited so far, since they are expected to yield good average performances.

Finally, it must be highlighted that the values of $p_{\theta_i}$ need to be recalculated at each iteration for all configurations, since the normalizing bounds can change across iterations.

\subsection{Regression Modeling}
\label{regmodmodsel}

The role of regression modeling in the proposed tuning framework is to enable predictions of the expected performance of a given configuration, conditional on its parameter values. For this, modeling strategies need ideally to be i) reasonably accurate; ii) capable of working with relatively few data points; iii) computationally inexpensive (at least relatively to the cost of evaluating the configurations); and iv) parsimonious in terms of the number of coefficients in the model. Another desirable trait is that the regression models scale reasonably well up to a reasonable number of parameters, e.g., $10$ (which is a reasonable upper limit for free parameters that are expected to be adjusted by the user).

For continuous parameters, usual models include linear regression with ordinary \cite{Montgomery2012} or weighted \cite{Strutz2010} least square estimators; quantile regression \cite{Koenker2000}; and ridge or lasso regression \cite{Tibshirani1996}, among others. In this work we opted for using the latter, which is briefly introduced below.

\subsubsection{Ridge and Lasso Regression}
\label{ridgelasso}

There are two reasons why ordinary least squares (OLS) regression is often inadequate \cite{Tibshirani1996}, namely prediction accuracy and interpretation. Poor prediction accuracy can be caused by a low bias and large variance of \textit{OLS} regression, whereas interpretation is often challenging given the large number of coefficients often used when regressing models with more than a couple of variables. 

As a way to remedy both these issues, some methods employ shrinkage techniques to remove coefficients that do not contribute to the explanatory power of a given model. Shrinking a coefficients can be achieved, e.g., by including a penalty term in the problem of minimizing the least squared errors. Considering the predictor of $p_{\theta_i}$ as a linear function of the form $\hat{p}_{\theta_i} = \beta_0 + \theta_i^T\beta:~\beta_0\in\mathbb{R},~\beta\in\mathbb{R}^p$, the problem becomes \cite{Owen2006}:
\begin{equation}
\mbox{Find}~~\beta^* = \argmin_{\beta\in\mathbb{R}^p} \sum\limits_{i=1}^{n}(p_{\theta_i} - \beta_0 - \theta_i^T\beta)^2 + \lambda \norm{\mbs{\beta}}_{\alpha}^2
\label{eq:ridgelasso}
\end{equation}

\noindent where $\lambda \in [0,\infty]$ is a regularization parameter, and $\alpha\in\mathbb{Z}_{>0}$ regulates the order of the norm used for the penalization term. Two special cases of this definition are the \textit{lasso} ($\alpha = 1$) and \textit{ridge} ($\alpha = 2$) regressions. The minimization of \eqref{eq:ridgelasso} becomes more aggressive at shrinking coefficients towards zero (i.e., removing their associated terms from the model) as larger values of $\lambda$ are used. This regression approach can be useful in the presence of complex models with many terms, particularly when there is a large difference in the relevance of each term, as is often the case of algorithm parameters \cite{Nannen2006,Nannen2007}. In these cases, shrinkage will reduce all coefficient values, leading those least relevant to zero and amplifying the differences between them, simplifying the model and facilitating interpretation.

\subsection{Generating Perturbed Models}
\label{genmodels}

The generation of response surfaces to be optimized at each iteration is performed in two steps: firstly, a regression model is fit using a modeling technique of choice. Secondly, the model obtained is perturbed several times, generating new response surfaces (line \ref{alg:perturbmodels} of Algorithm \ref{algo:metatun}). To generate the perturbed models, all (non-zero) coefficients of the model are subject to uniform noise. The range of this noise is defined by the standard errors of each coefficient, which can be obtained either analytically (e.g., in the case of linear regression models using OLS) or by resampling methods. 

For ridge and lasso regression models, standard errors are obtained using a leave-one-out (LOO) strategy. After fitting a model using the approach described in the previous section, all coefficients that were shrunk down to zero are removed from the model. The resulting polynomial is then used as a basis for fitting $k$ new models, each of which is fitted on a dataset obtained by ignoring the information regarding a single configuration. Based on the coefficients fit for each of these $k$ leave-one-out models,  the standard error of each coefficient is estimated as the sample standard deviation of the values obtained for that coefficient on all LOO models.

\subsection{Model Optimization}
\label{sec:modeloptim}

Considering that the response surfaces represent preliminary models of the  average behavior of the algorithm conditional on its parameter values, optimizing them should yield a set of new candidate configurations with expected good performance. As the iterations progress and more candidate configurations are evaluated on more instances, it is expected that the resulting models become increasingly accurate.

The main concept of the proposed tuning framework is to iteratively fit regression models of average behavior as a function of parameter values, using increasing amounts of information, and optimizing the resulting response surfaces (and perturbed versions of them, obtained by incorporating estimation uncertainties of the model coefficients - lines \ref{alg:optim1}--\ref{alg:optim2} of Algorithm \ref{algo:metatun}) to search for more promising parameter values. The new candidate configurations returned by optimizing these models are evaluated on all instances already visited by the method (line \ref{alg:evalcand3}) and added to the archive (lines \ref{alg:arch1}--\ref{alg:arch2}).

The optimization approach to be used depends on the nature of the regression models, which may provide, e.g., analytical gradients or guarantees of unimodality. For more general or complex models, fast heuristics can be employed. In this work we opted for using Nelder-Mead Simplex \cite{Nelder1965,Nash2017} to optimize the response surfaces.

\section{Experimental Results}
\label{sec:results}

To illustrate the use of the proposed approach, we performed the tuning of six parameters of the MOEA/D algorithm \cite{journal.Zhang2007,report.Campelo2017} for the hypothetical problem class represented by the 2-objective problems of the UF benchmark set \cite{report.Zhang2008,journal.Bossek2017}, with dimensions between 3 and 40. Dimensions that are multiples of $3$ were reserved for testing (91 test instances), while all others (175 instances) were available for the tuning effort. The quality of the solutions returned by the algorithm on each problem instance was computed using the IGD indicator \cite{Zitzler2000}.

The complete specification of the MOEA/D algorithm and its fixed and tunable parameters is shown in Table \ref{tab:moead}. A detailed description of the MOEA/D algorithm and its component-based modeling can be found in the available literature \cite{journal.Zhang2007,arxiv.MOEADr}.

\renewcommand{\arraystretch}{1.3}
\begin{table*}[htb]
	\centering
	\begin{tabular}{|c|c|c|}
		\hline
		\textbf{Module} & \textbf{Type} & \textbf{Parameter (value / range)}\\
		\hline
		Decomposition method& SLD& $h = 99$\\
		\hline
		Scalar aggregation function& PBI& $\theta^{pbi}\in\left[0,10\right]$\\
		\hline
		Objective scaling& None & --\\
		\hline
		Neighborhood assignment& In the variable space &$\delta_p\in\left[0.5,1\right]$\\
		\hline
		\multirow{5}{*}{Variation Operators} & \multirow{2}{*}{SBX recombination}& $\eta_{\mathtt{X}}\in\left[0.5, 100\right]$\\
		& & $p_{\mathtt{X}}\in\left[0.25,1\right]$\\
		\cline{2-3}
		& \multirow{2}{*}{Polynomial mutation}& $\eta_{\mathtt{M}}\in\left[0.5, 100\right]$\\
		& & $p_{\mathtt{M}} = 1 / dim$\\
		\cline{2-3}
		& Simple truncation & --\\
		\hline
		Update strategy& Restricted& $n_r\in\left[1,20\right]$\\
		\hline
		Termination criteria& Function evaluations& $max_{eval} = 1000\times dim$\\
		\hline
	\end{tabular} 
	\caption{MOEA/D configuration used in the experiments. Tunable parameters are defined in ranges. $dim$ is the dimension of the instance being solved.}
	\label{tab:moead}
\end{table*}

For this tuning problem, the proposed framework was set up as follows: the initial sampling was performed with $m_0 = 100$ configurations, randomly generated using Latin Hypercube sampling and evaluated using $n_0 = 5$ instances randomly drawn from the tuning set. At each iteration, $m_i = 10$ new configurations were generated, and $n_i = 1$ new tuning instance was added to the pool. The summary function used for calculating the expected performance of each configuration was the median. Regression models were fit using Lasso regression, using a polynomial model of order $4$ as the basis.\footnote{The penalization parameter $\lambda$ was selected automatically by cross validation. All models were fit using the implementation available in the \texttt{R} package \texttt{hqreg} \cite{hqreg}.} Perturbed models were generated using standard errors obtained by leave-one-out resampling, and the optimization of response surfaces was performed using the Nelder-Mead Simplex algorithm. A computational budget of $1000$ runs was defined for the tuning effort.

To obtain a comparison baseline, we used the official implementation of Iterated Racing, available in \texttt{R} package \texttt{irace} \cite{Lopez-Ibanez2016}. The standard configuration was used, and the same computational budget was imposed. Thirty repetitions of the tuning effort were performed using the set of tuning instances defined earlier. For each repetition, the best configuration returned was used to solve the instances in the test set, and its performance on each test instance was calculated as the mean IGD of 14 runs. The resulting observations were used to compare the performance of the two parameter tuning approaches.

The proposed approach and the Iterated Racing were then compared using a paired t-test \cite{Montgomery2012} at a $95\%$ significance level,\footnote{The test assumptions were validated using graphical analysis of the residuals.} with \textit{instance} as the pairing variable (repeated runs were mean-aggregated to prevent pseudoreplication). No significant differences in mean performance were found ($t = -0.18,~df = 90,~p = 0.86$; $CI_{0.95} = -0.0005\pm0.0065$).  This test indicates not only the lack of statistically significant differences in mean performance, but also suggests (given the width of the resulting confidence interval) that this experiment had adequate statistical power to detect any differences greater than about $0.013$ in mean IGD \cite{arxiv.Campelo2018}.
\begin{figure}[ht]
	\centering
	\includegraphics[width=\linewidth]{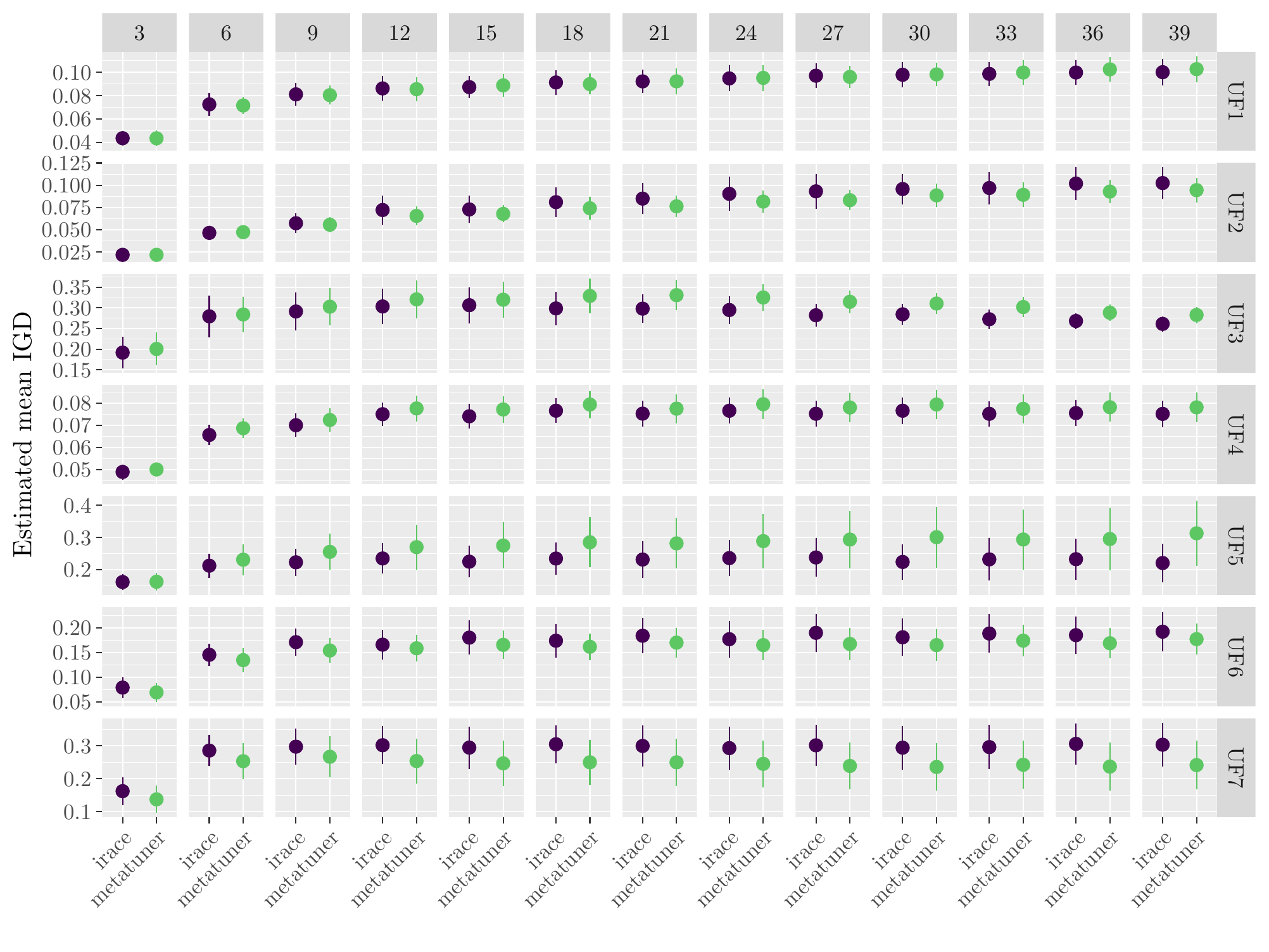}
	\caption{Comparison of mean IGD obtained by configurations returned by the proposed approach (light green) and \textit{irace} (dark purple) for the test instances. Vertical bars illustrate standard errors, not confidence intervals.}
	\label{fig:meanIGD}
\end{figure}

Figure \ref{fig:meanIGD} illustrates the estimated mean (with standard error bars) IGD values obtained by the configurations returned by each methodology on each instance. As expected, the performance of configurations returned by both methods tends to degrade as the problem dimension is increased, but both tend to follow similar patterns in all cases, which is reflected by the absence of statistically significant differences.

\begin{figure}[ht]
	\centering
	\includegraphics[width=\linewidth]{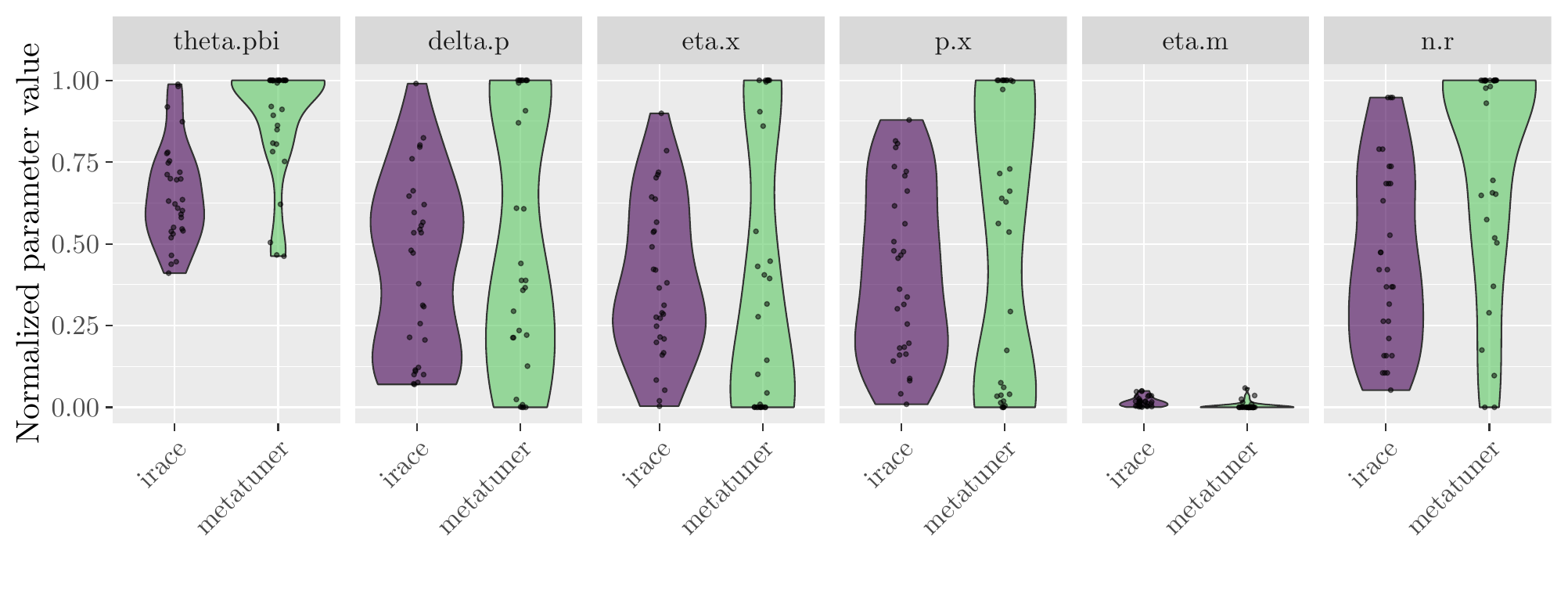}
	\caption{Variability of configurations returned by both methods. Parameter values were scaled to the range $[0,1]$ using the limits of the search range described in Table \ref{tab:moead}.}
	\label{fig:parameterVals}
\end{figure}

Figure \ref{fig:parameterVals} shows the variability of the parameter values (with the tuning range normalized to the interval $\left[0,1\right]$) of the best configurations returned by the tuning methods. Parameters $\delta_p$, $\eta_{\mathtt{X}}$ ,$p_{\mathtt{X}}$ and $n_{r}$ show a large variability within the search range defined in the tuning effort, which suggests a relative insensitivity of the MOEA/D performance to the values selected for these parameters, at least in terms of main (marginal) effects. Parameter $\theta^{pbi}$ shows a bias towards the upper half of the search range (i.e., values greater than $\approx 5$), while the polynomial mutation parameter $\eta_{\mathtt{M}}$ shows a very strong tendency to assume values very close to the lower bound of the tuning range (i.e., values close to $0.5$), which suggests a strong effect of this parameter on the IGD performance of the MOEA/D, for the problem class used in this example. It is interesting to observe that both the proposed method and the Iterate Racing exhibit the same general trends in terms of the variability in the parameter values returned.

While no statistically significant differences have been found between the proposed method and Iterated Racing, one advantage of the approach presented in this work is the explicit statistical modeling of the expected performance of the algorithm, as a function of its parameter values, which allows us to investigate the conjectures raised by the analysis of Figure \ref{fig:parameterVals}. Arbitrarily selecting the results of the first replicate of the proposed method as a representative example,\footnote{For reference, the configuration returned by this particular execution was $\theta^{pbi} = 9.92$, $\delta_p = 0.954$, $\eta_{\mathtt{X}} = 0.600$, $p_\mathtt{X} = 0.260$, $\eta_{\mathtt{M}} = 0.500$, and $n_r = 4$.} the lasso regression (which automatically shrinks coefficients with very small effects towards zero) returned a model of the expected IGD performance of the MOEA/D on the problem class represented by the tuning instances as:
\begin{equation}
\label{eq:regression}
\begin{split}
y = 0.226 - 6.67\times 10^{-3}\theta^{pbi} &- 8.86\times 10^{-3}\theta^{pbi}\delta_p +\\
				 & +6.04\times 10^{-4}\eta_{\mathtt{M}} +  1.20\times 10^{-5}\eta_{\mathtt{M}}\eta_{\mathtt{X}}
\end{split}
\end{equation}

\noindent where $y$ represents the expected IGD of the algorithm on the problem class represented by the tuning instances. This result not only highlights the importance of the parameters identified in Figure \ref{fig:parameterVals} as possibly relevant, i.e., $\theta^{pbi}$ and $\eta_{\mathtt{M}}$, quantifying the effects of increasing / decreasing these parameters on the expected IGD performance; but also suggests two others ($\delta_p$ and $\eta_{\mathtt{X}}$) which, while not critical in terms of their main effects, present contributions to the performance of the algorithm in terms of interactions with the two most relevant parameters.

\begin{figure}[htb]
	\centering
	\includegraphics[width=\linewidth]{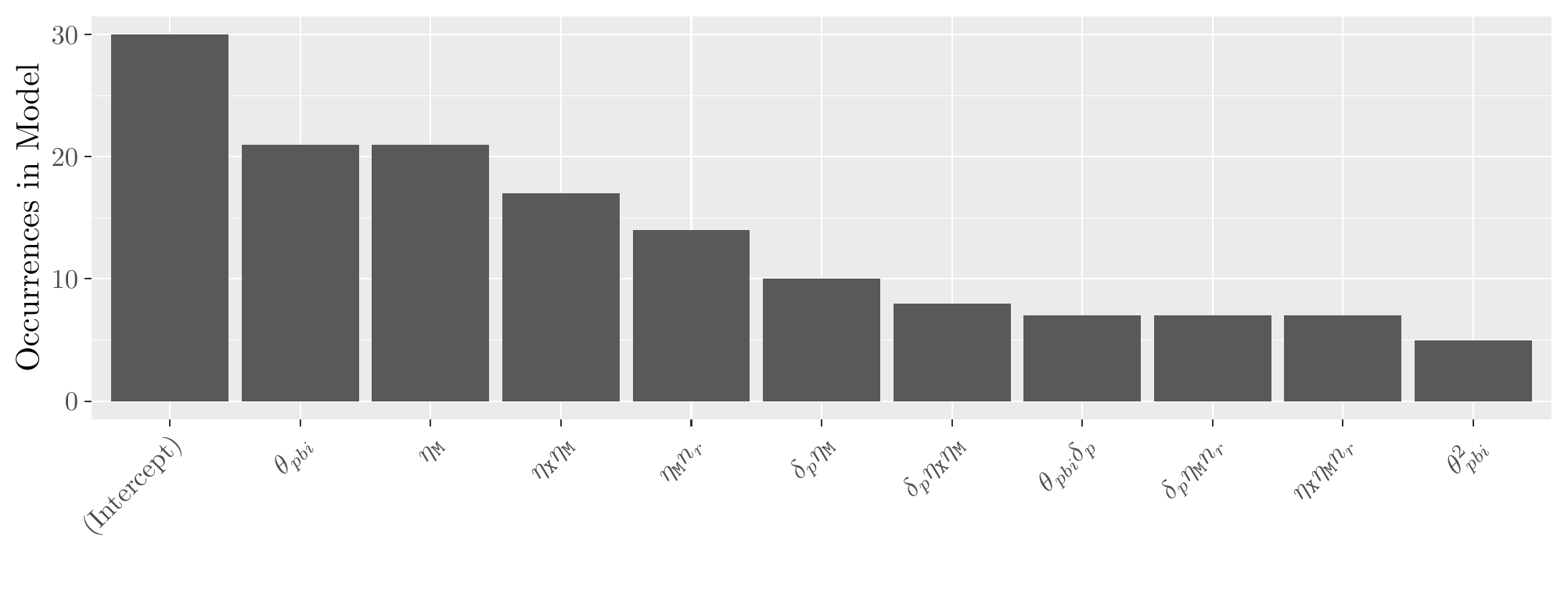}
	\includegraphics[width=\linewidth]{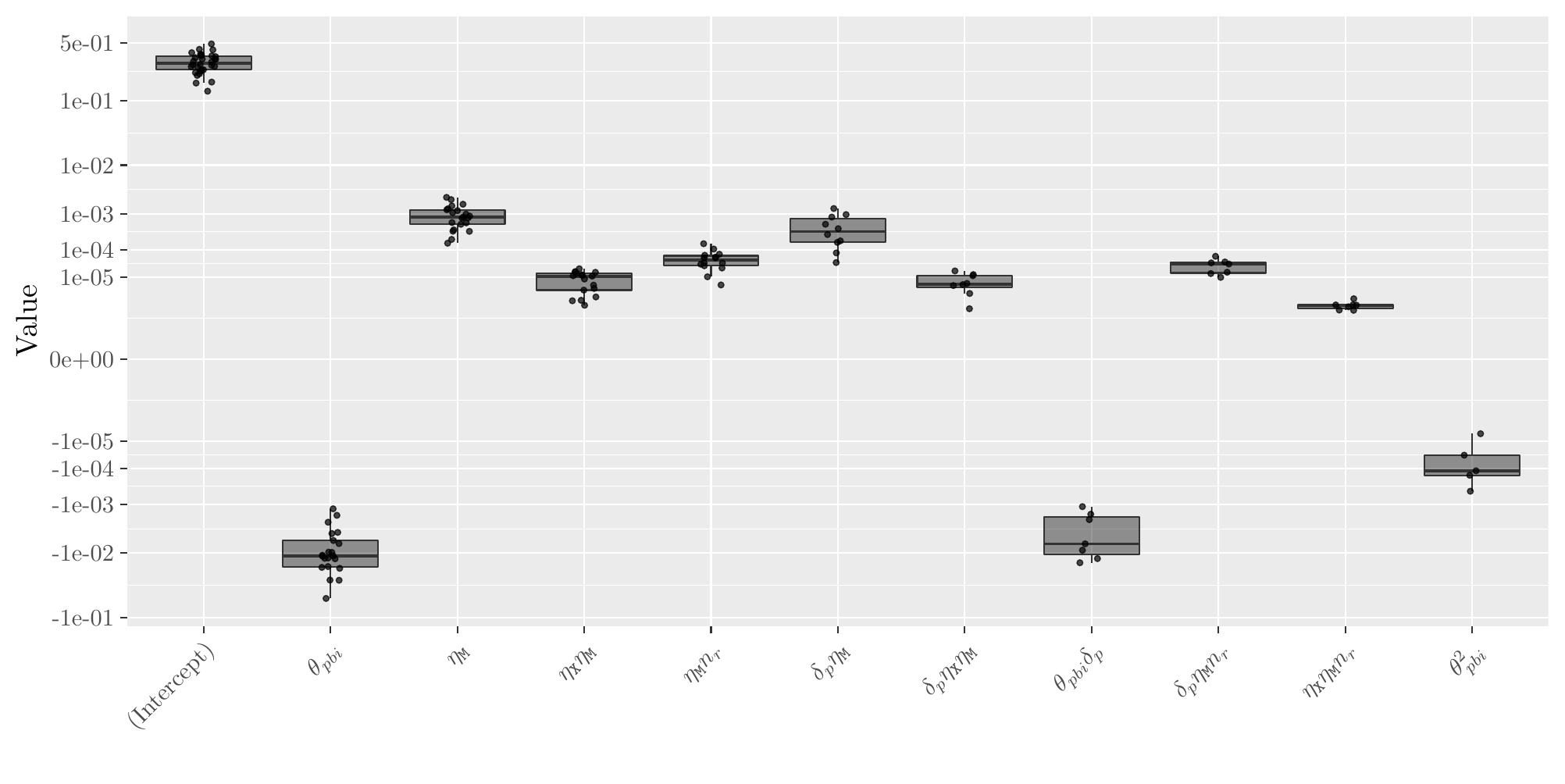}
	\caption{Non-zero coefficients commonly returned by the regression modeling. \textbf{Top}: number of times each term had a non-zero coefficient in the models output by the proposed methodology, from a sample of 30 replicates. Only terms that had non-zero coefficients five or more times are shown. \textbf{Bottom}: distribution of non-zero values returned for each coefficient. Notice that the vertical scale is compressed.}
	\label{fig:coef_occurrences}
\end{figure}

Finally, since the lasso regression \cite{Tibshirani1996,hqreg} tends to shrink coefficients towards zero based on the explanatory value of their respective terms (which is conditional on the sample used for fitting the model), it can be interesting to investigate which terms were generally preserved across repeated runs of the method. Figure \ref{fig:coef_occurrences} illustrates the model terms that appeared more frequently in the models returned by the proposed tuning method, as well as their usual (non-zero) values. 

\clearpage

From this figure it is clear that the main effects of $\theta^{pbi}$ and $\eta_{\mathtt{M}}$ were generally detected by the modeling approach as being particularly relevant, as well as $\eta_{\mathtt{M}}\eta_{\mathtt{X}}$ and a few other interaction terms. These results point at some interesting insights regarding the influence of these parameters on the performance of the MOEA/D. For instance, larger values of $\theta^{pbi}$ (which are associated with the selection bias of the MOEA/D towards greater diversity on the space of objectives), tend to predict a significant improvement (i.e., reduction) in terms of the expected IGD. Further parametric analyses can also be performed, but this detailed exploration falls outside the scope of the current paper.

\section{Conclusions}
\label{sec:conclusions}

In this work we present a new parameter tuning framework based on concepts from Sequential Model Based Optimization (SMBO) methods. The proposed framework is centered on the sequential optimization of perturbed regression models of expected algorithm performance conditional on parameter values, and on the sequential evaluation of new problem instances on the most promising candidate configurations.

The proposed framework was instantiated into a method for tuning numeric parameters, and a case study was presented using the Iterated Racing approach as a comparison baseline. The results suggest that, in terms of raw performance, the proposed framework was able to return configurations that matched the performance of iterated racing, which is generally considered a particularly efficient tuning approach. Moreover, the proposed method is also designed to provide insights on the relevance of  the parameters being tuned, as well as predictive models of expected performance conditional on the parameter values, which can inform further development of the algorithm or the definition of standard values for practical implementations, or suggest 

While the results presented in this work suggest that the proposed approach may be an interesting alternative for parameter tuning, and possibly for the investigation of design aspects in algorithm research, further testing and development are needed to effectively establish its power and limitations. For instance, while both the lasso regression and the Nelder-Mead Simplex algorithm used in the tuning method presented in this work can be effectively used for high-dimension situations, the limitations of the proposed framework in terms of the number of parameters that can be tuned remains to be investigated. Moreover, the adaptation of the principles described here to the tuning of categorical or hierarchical parameters is also a field which we have not yet investigated, and represent possibilities of continuity for this work.

\newpage
\bibliographystyle{spmpsci}      % mathematics and physical sciences
%\bibliography{00_SWRM2018-TrindadeCampelo}   % name your BibTeX data base

\end{document}